\documentclass{article}
\RequirePackage[loading]{tracefnt}

\PassOptionsToPackage{numbers, compress}{natbib}

\usepackage[final]{nips_2018}

\usepackage[utf8]{inputenc} 
\usepackage[T1]{fontenc}    
\usepackage{hyperref}       
\usepackage{url}            
\usepackage{booktabs}       
\usepackage{amsfonts}       
\usepackage{nicefrac}       
\usepackage{microtype}      

\usepackage{graphicx}
\usepackage{epstopdf}
\title{Modeling disease progression in longitudinal EHR data using continuous-time hidden Markov models}

\author{
  Aman Verma \\
  Department of Epidemiology\\
  McGill University\\
  \texttt{aman.verma.mtl@gmail.com} \\
  \And
  Guido Powell \\
  Department of Epidemiology\\
  McGill University\\
  \And
  Yu Luo \\
  Department of Biostatistics\\
  McGill University\\
  \And
  David Stephens \\
  Department of Mathematics and Statistics\\
  McGill University\\
  \And
  David L. Buckeridge \\
  Department of Epidemiology\\
  McGill University\\
}

\begin{document}

\maketitle

\begin{abstract}
Modeling disease progression in healthcare administrative databases is complicated by the fact that patients are observed only at irregular intervals when they seek healthcare services. In a longitudinal cohort of 76,888 patients with chronic obstructive pulmonary disease (COPD), we used a continuous-time hidden Markov model with a generalized linear model to model healthcare utilization events. We found that the fitted model provides interpretable results suitable for summarization and hypothesis generation.
\end{abstract}

\section{Introduction}
Chronic diseases typically progress slowly over many years. For example, patients with chronic obstructive pulmonary disease (COPD)  may progress from mild to very severe stages of the disease over the span of more than a decade \cite{gold_executive_committee_pocket_2015}. By advancing our ability to model such disease progression, we can improve early detection, proper management and accurate prognosis of diseases. Crucial to modeling disease progression is making use of the vast amounts of individual patient data in the form of longitudinal health records, such as electronic health records (EHRs) and healthcare administrative databases. However, processes like chronic disease progression are not observed directly or explicitly recorded in these databases, but are rather underlying processes that generate the observations. As a result, inferential methods are needed to identify how patients move through stages of a disease. 

Adding to the complexity of modeling disease progression is the fact that patient data is only recorded when they receive services from the health system, resulting in very irregularly-spaced observations that will differ in granularity between patients and across time within a patient. Disease trajectories, in terms of rates of progression and profiles of healthcare use, will also vary widely across patients, further creating challenges in synthesizing progression. Furthermore, for large segments of patient trajectories records may be very sparse due to infrequent visit or in many cases incomplete (e.g., censored data or care provided outside the health system). 

To deal with the heterogeneous, sparse, non-equidistant, and incomplete longitudinal observations in a cohort COPD patients we apply our recently developed continuous-time hidden Markov model (CTHMM) under a generalized linear modeling framework \cite{luo_latent_nodate}. Such multi-state models can capture patient status over time as a discrete-time realization of a continuous-time Markov process, while accommodating irregular spacing of observations \cite{kalbfleisch_analysis_1985}. Previous research in disease progression modeling has typically made use of known disease status or transition rates \cite{andersen_multi-state_2002, meira-machado_multi-state_2009, lange_joint_2015} rather than modeling the latent process as a stochastic process. By using a hidden Markov model (HMM), we assume that the Markov property is imposed on the unobserved sequence governing the observations. In this study, we fit a CTHMM to a large cohort of COPD patients to evaluate how well the model could generate hypotheses about the evolution of the illness and how to improve its management.

\begin{figure}
  \centering
  \includegraphics[width=\linewidth]{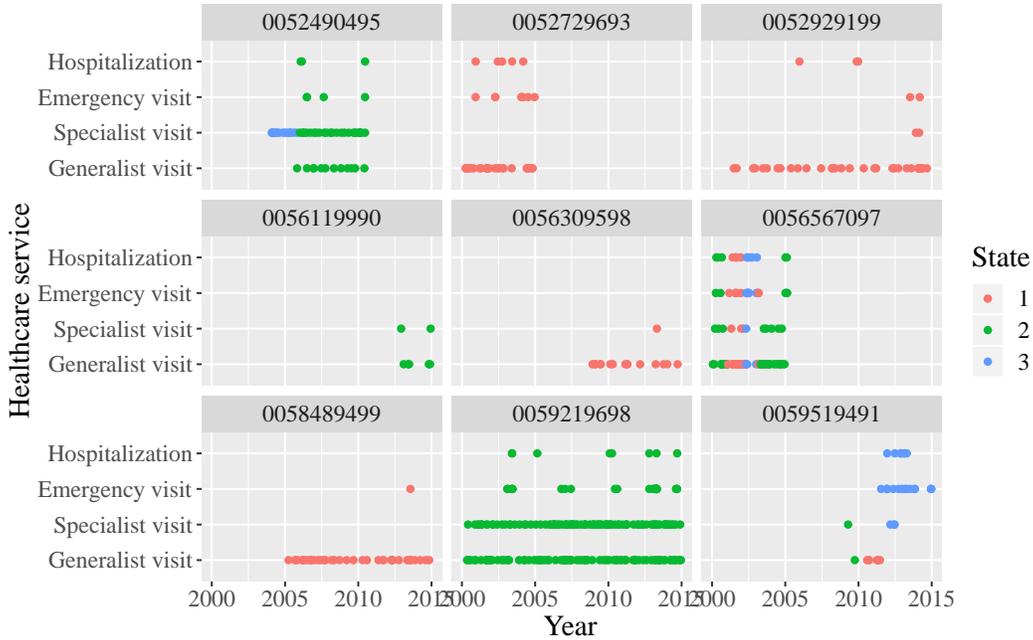} 
  \caption{Healthcare service by time with most probable hidden state for nine sample patients.}
\end{figure}

\section{Methods}
\subsection{Data}
A cohort of COPD patients was selected from an open, dynamic cohort beginning in 1998 (with yearly re-sampling), capturing a 25\% random sample drawn from the registered with the provincial health insurance agency (\emph{Régie de l'assurance maladie du Québec}) in the census metropolitan area of Montreal. Patients were followed until they died or moved out of the region. The administrative records include outpatient diagnoses and procedures submitted through billing claims, and procedures and diagnoses from hospital records. 

Using established case-definitions based on diagnostic codes \cite{blais_quebec_2014}, we enrolled 76,888 COPD patients as of their incident event occurring after a minimum of two years of time at risk (ICD-9 491x, 492x, 496x; ICD-10 J41-J44) with follow-up until December 2014. When patients were observed by the health system, their discrete state was only indirectly measured through a proxy of types of health services received, specifically, outpatient visits with a general practitioner (GP), outpatient visits with a specialist (Spec) (coded as respirologist or internist), visits to an emergency department (ED), and all-cause hospitalizations (Hosp). Health service uses were not considered if they occurred prior to the incident COPD event.

\subsection{Model description}
To model disease progression within the COPD cohort, we used a continuous-time hidden Markov model (CTHMM). The observable data in this CTHMM was the healthcare utilization event, which was one of four mutually exclusive events (GP, Spec, ED, Hosp). If events occured on the same day, then we used the most severe event, where Hosp was the most severe, follwed by ED, Spec, and GP.

Given that there was an observed event, the probability of each of the four observables was determined by a multinomial model, using GP as the reference event. The multinomial model had a set of three binary covariates, coding one of four previous observed events, along with an intercept.

Because the observed healthcare utilization events were irregularly-spaced in time, we modeled the transitions between hidden states in continuous time. Because patients may have extended periods without any healthcare visits, we allowed for transitions between states even if there were no observed healthcare utilization events. The transition rate between states was specified by a rate matrix $Q$, which was fit as part of the model. The hidden states were interpreted as belonging to states of disease severity, by computing the probability of healthcare utilization events with known levels of severity.

\begin{figure}
  \centering
  \includegraphics[width=\linewidth]{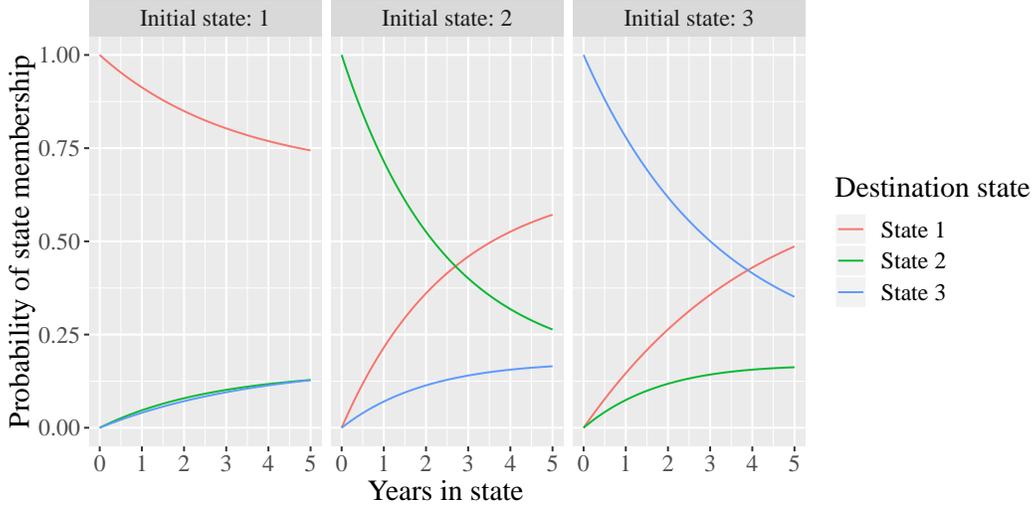} 
  \caption{Probability of state membership by time for each of three starting states.}
\end{figure}

\subsection{Model fit}
Our model required us to fit three sets of parameters. $\pi$ specified the probability of starting in each state, $Q$ specified the transition matrix between each state, and $\beta$ specified the probability of each observable for each state. 

For each state, there was a set of parameters $\beta$ that specified the multinomial model of the four healthcare utilization outcomes (GP, specialist, ED, hospital). Each state's multinomial model was specified as a set of three logistic regression models of the probability of specialist, ED, or hospital visit versus the probability of a GP visit. Each logistic regression model had four parameters: an intercept, and the previously observed healthcare utilization visit specified as three binary variables (using one-hot encoding). With four parameters within each of three logistic regression models within each of three states, we had 36 $\beta$ parameters.

The transition rate between states was specified by the $3 \times 3$ (each row and column corresponding to a state) rate matrix $Q$. If $X_t$ is the state membership at time $t$, then  $Q_{ij} = \lim_{\Delta t \to 0} \frac{P(X_{t + \Delta t} | X_t)}{\Delta t}$ where $i \neq j$. The diagonal entries of $Q$ are computed so that the rows of $Q$ sum to zero. Since only the non-diagonal entries need to be fit, there were six $Q$ parameters in our model. We scaled $\Delta t$ to be expressed in years.

To fit our model, we used expectation-maximization (EM) \cite{dempster_maximum_1977}, an iterative algorithm in which previous parameter values are used to compute new values. EM equired us to specify starting values for the model parameters. For $\pi$, we specified equal probability of starting in any state, the $\beta$ parameters were randomly drawn from a normal distribution with a mean of 0 and a standard deviation of 1, and the $Q$ parameters were drawn from a uniform distribution between 0 and 1.

At each iteration of EM, we used the forward-backward algorithm \cite{baum_inequality_1967, baum_growth_1968}, along with the current parameter values for $\pi$, $\beta$, and $Q$, to compute, for each consecutive pair of observations within each patient, the probability of starting and ending in each state pair. For each consecutive pair of observations, we computed a $3 \times 3$ matrix of probabilities that summed to 1. We used these matrices to also compute the marginal probability of state membership at all observed times for all patients.

We computed a new value for $\pi$ using the mean probability of state membership at the first observed time point across all patients, and computed new values for $\beta$ by fitting a multinomial model for each state, weighting each observation by the probability of state membership at that time. Finally, we computed new values for $Q$ by using a nested EM procedure \cite{bladt_statistical_2005, metzner_generator_2007}. We stopped the algorithm when the difference in norms between the previous and new parameter values was below 0.05.

\section{Results}

Table \ref{tab:prob} describes the $\beta$ coefficients of a 3-state model (a 4-state model resulted in a less meaningful "transient" state to which the probability of transitioning was persistently minimal, i.e. < 0.10). The $\beta$ coefficients can be compared at the autoregressive term of GP visits since it is the most frequent health service. State 1 mainly represents a well-controlled state with high probability of GP visits (0.89), state 2 represents a mix of GP visits (0.53) and specialist visits (0.34), while state 3 represents a more severe state where ED visits (0.40) and hospitalizations (0.18) are more likely than in the other states. The $\beta$ coefficients distributions are mostly similar when compared at other autoregressive terms. However, at the specialist visits autoregressive term in the more "severe" state 3, the most probable event is specialist visits (0.65), with a lower probability of acute care, suggesting differences in how disease severity progresses for patients consulting a specialist.

After 5 years, patients are more likely to remain in state 1 (0.74) or transition to it from states 2 (0.57) or 3 (0.48). While 5-year transitions to states 2 and 3 are less frequent (ranging from 0.13 to 0.16), patients have a higher probability of remaining in state 3 after 5 years (0.35) than remaining in state 2 (0.26), suggesting a less favourable rate of recovery from the more severe state. 

\begin{table}
  \caption{Probability of each healthcare utilization event, given current state, and last observed healthcare utilization event. The rows stratify the last observed event, while the columns stratify the probability of each new event, by state. GP = general practition visit, ED = emergency department visit,  Hosp = hospitalization, Spec = specialist vist.}
  \label{tab:prob}
  \centering
  \resizebox{\columnwidth}{!}{%
  \begin{tabular}{lrrrrrrrrrrrrrr}
    \toprule
    \multicolumn{1}{c}{} & \multicolumn{4}{c}{State 1} & \multicolumn{4}{c}{State 2} & \multicolumn{4}{c}{State 3}
    \\
    \cmidrule(r){2-5}
    \cmidrule(r){6-9}
    \cmidrule(r){10-13}
    & GP      & ED     & Hosp & Spec
    & GP      & ED     & Hosp & Spec
    & GP      & ED     & Hosp & Spec  \\
    \midrule
GP & 0.89 & 0.07 & 0.02 & 0.01 & 0.53 & 0.09 & 0.04 & 0.34 & 0.33 & 0.40 & 0.18 & 0.09 \\
ED & 0.50 & 0.32 & 0.17 & 0.01 & 0.30 & 0.32 & 0.2 & 0.18 & 0.11 & 0.57 & 0.26 & 0.06 \\
Hosp & 0.71 & 0.15 & 0.12 & 0.02 & 0.44 & 0.16 & 0.11 & 0.29 & 0.20 & 0.48 & 0.17 & 0.15 \\
Spec & 0.78 & 0.05 & 0.10 & 0.07 & 0.55 & 0.07 & 0.04 & 0.35 & 0.08 & 0.19 & 0.07 & 0.65 \\
    \bottomrule
  \end{tabular}%
  }
\end{table}

\section{Discussion}

Using a continuous-time hidden Markov model, we used healthcare utilization within a cohort of COPD patients as a proxy for unobserved, latent, disease progression. We found that the model provided interpretable results, allowing us to generate hypotheses about how healthcare utilization evolves at different stages of disease progression. The model allowed us to learn dynamic changes in underlying disease state from heterogeneous, sparse and incomplete observations. The discrete states suggest increasing levels of severity based on a progression away from mostly primary care, to more specialist care, and finally more acute care. 

By modeling complex longitudinal observations into latent disease states, we have created opportunities for future research in understanding COPD. The CTHMM can include time-varying covariates such age, comorbidities, prescribed drugs, or socioeconomic status to identify important predictors of progression towards greater severity. Such models can improve prognoses by matching patients to typical trajectories of a sub-population. Also, decision makers can identify health service profiles that are predictive of slower disease progression. 

\medskip

\small

\bibliography{nips_2018}

\end{document}